\documentclass[twocolumn, switch]{article} 

\usepackage{preprint}

\usepackage{amsmath, amsthm, amssymb, amsfonts}
\usepackage{amstext}
\usepackage{mathrsfs} 

\usepackage[numbers,square]{natbib}
\usepackage[utf8]{inputenc}	
\usepackage[T1]{fontenc}	
\usepackage{xcolor}		
\usepackage[colorlinks = true,
            linkcolor = purple,
            urlcolor  = blue,
            citecolor = cyan,
            anchorcolor = black]{hyperref}	
\usepackage{booktabs} 		
\usepackage{nicefrac}		
\usepackage{microtype}		
\usepackage{lineno}		
\usepackage{float}			

\usepackage{lipsum}		

\usepackage{newfloat}
\DeclareFloatingEnvironment[name={Supplementary Figure}]{suppfigure}
\usepackage{sidecap}
\sidecaptionvpos{figure}{c}

\usepackage{titlesec}
\titlespacing\section{0pt}{12pt plus 3pt minus 3pt}{1pt plus 1pt minus 1pt}
\titlespacing\subsection{0pt}{10pt plus 3pt minus 3pt}{1pt plus 1pt minus 1pt}
\titlespacing\subsubsection{0pt}{8pt plus 3pt minus 3pt}{1pt plus 1pt minus 1pt}

\usepackage{tikz,xcolor,hyperref}

\definecolor{lime}{HTML}{A6CE39}
\DeclareRobustCommand{\orcidicon}{
	\begin{tikzpicture}
	\draw[lime, fill=lime] (0,0) 
	circle [radius=0.16] 
	node[white] {{\fontfamily{qag}\selectfont \tiny ID}};
	\draw[white, fill=white] (-0.0625,0.095) 
	circle [radius=0.007];
	\end{tikzpicture}
	\hspace{-2mm}
}
\foreach \x in {A, ..., Z}{\expandafter\xdef\csname orcid\x\endcsname{\noexpand\href{https://orcid.org/\csname orcidauthor\x\endcsname}
			{\noexpand\orcidicon}}
}

\title{Multitask Multi-database Emotion Recognition}

\usepackage{xwatermark}

\usepackage{authblk}

\author[1,2]{Manh Tu VU}
\author[2]{Marie Beurton-Aimar}

\affil[1]{Lucine, 223 Avenue Emile Counord, 33300 Bordeaux, France}
\affil[2]{LaBRI, 351 Cours de la Libération, 33405 Talence CEDEX, France}

\begin{document}

\twocolumn[ 
  \begin{@twocolumnfalse} 
  
\maketitle

\begin{abstract}
  In this work, we introduce our submission to the 2nd Affective Behavior Analysis in-the-wild (ABAW) 2021 competition. 
  We train a unified deep learning model on multi-databases 
  to perform two tasks: seven basic facial expressions prediction and
  valence-arousal estimation. Since these databases do not contain
  labels for all the two tasks, we have applied the distillation
  knowledge technique to train two networks: one teacher and one
  student model. The student model will be trained using both ground
  truth labels and soft labels derived from the pretrained teacher
  model. During the training, we have added one more task, which is the
  combination of the two mentioned tasks,  
  for better exploiting inter-task correlations. We also exploit the
  sharing videos between the two tasks of the AffWild2 database  
  that is used in the competition, to further improve the performance
  of the network. Experiment results show that the network has
  achieved promising results on the validation set of the AffWild2
  database.  
  Code and pretrained model are publicly 
  available at \url{https://github.com/glmanhtu/multitask-abaw-2021}
\end{abstract}
\vspace{0.35cm}

  \end{@twocolumnfalse} 
] 



\section{Introduction}
Emotion recognition and analysis is a crucial part of many applications and systems, especially in health care 
and medical fields \cite{Thevenot.2017.2754861, pain_survey} as it is directly related to the health state of a patient. 
As results, more and more works have been conducted to try to analyse human 
emotions and behaviours \cite{sebe2005multimodal, saxena2020emotion, Werner_Lop}. In the same sense, the 2nd Affective Behavior 
Analysis in-the-wild (ABAW 2021) competition by Kollias et al. \cite{ 
kollias2020analysing, kollias2021distribution, kollias2021affect, kollias2019expression, kollias2019face, 
kollias2019deep, zafeiriou2017aff, kollias2021analysing} provides a large-scale dataset
Aff-Wild2 for analysing human emotion in-the-wild settings. This
dataset includes annotations for three tasks, which are including: 
valence-arousal estimation, action unit (AU) detection, and seven basic facial expression classification.
Valence represents how positive the person is while arousal describes how active this person is. 
The seven basic facial expressions include neutral, anger, disgust, fear, happiness, sadness, and surprise.
AUs are the basic actions of individuals or groups of muscles for portraying emotions. 

In this paper, we focus on two tasks, which are including seven basic facial expressions classification and 
valence-arousal estimation. Inspired by the multitask training with incomplete label method from Deng et al. \cite{Deng}
we propose a method to futher exploit the inter-task correlations
between the two tasks. Similar to Deng et al. \cite{Deng} 
we also apply the distillation knowledge technique to train two multitask models: a teacher model and a student model.
However, instead of treating each task independently when training teacher model as in \cite{Deng}, we add one 
more task to the training process, which is the combination of the two tasks above to train the network 
using data comming from AffectNet database \cite{Mollahosseini_2019}, in which contains labels for both the two tasks. 
A part from that, we also try to exploit the shared videos (videos
annotated both seven basic facial expression and valence-arousal
labels) in the Affwild2 database by integrating this information to the
student model's training process (see Equation \ref{eq:student_if}). 
With these improvements, our model achieves a competitive results on
the validation set of the official dataset Affwild2 of the competition.

\section{Methodology}
In this section we introduce our multitask multi-databases method. 
Visual images are extracted from video and fed into a ResNet 50
\cite{he2015deep} networks to train for analysing human's emotion
in-the-wild. Then, features extracted from this network will go
through a GRU \cite{cho2014learning} network to capture temporal
information and finally, perform both the seven basic facial
expressions classification and valence-arousal estimation. Because in
our dataset, we don't always have all labels for all of our
tasks. Therefore, we have applied the multitask training with missing
labels method that is described in \cite{Deng} with some enhancements,
which we will describe in the sections below.

\subsection{Data Imbalancing}
Similar to \cite{Deng}, we also use some external datasets to address
the data imbalance problem in the Affwild2 dataset
\cite{kollias2019affwild4}. The external datasets are including
Expression in-the-Wild (ExpW) dataset \cite{zhang2018facial} for
expression classification and AFEW-VA dataset \cite{kossaifi2017afew}
for valence-arousal estimation. After merging these datasets, we 
have also applied the same dataset balancing protocol, which is described in \cite{Deng} to improve the balance of the
dataset.

Different from \cite{Deng}, we perform only two tasks: seven basic
facial expressions prediction and valence-arousal estimation.  
Since detecting AUs is out of our interest and we would like to perform the two mentioned tasks as best as possible.
we exclude the task of facial AU detection out of our training process. 
A part from that, we also want to include the AffectNet database \cite{Mollahosseini_2019} into the training, 
since in this database annotation for  both seven basic expressions
and valence-arousal are available. Now, for the training process, our dataset is including three parts:

\paragraph{Mixed EXPR} The mixing set of the AffWild 2 (expressions part) and ExpW datasets for seven basic expressions. 
This dataset has no information about valence and arousal.

\paragraph{Mixed VA} The mixing set of the AffWild 2 (valence-arousal part) and AFEW-VA datasets for valence and arousal.
This dataset has no information about the seven basic expressions.

\paragraph{Affect EXPR\_VA} The AffectNet dataset, for both seven basic expressions and valence-arousal.

Corresponding to these three dataset's parts are the three training tasks $\mathscr{T} \in \{1, 2, 3\}$, which are including: expression classification (EXPR), 
valence-arousal estimation (VA) and the mixed of these two tasks (EXPR\_VA). Since the data for the last task EXPR\_VA 
has annotated for both seven basic expressions and valence-arousal, this task will play the role 
of guiding the training, i.e. re-balancing the gradient back propagation for the first two tasks and exploiting 
the inter-task correlations. One can note that even though we have three training tasks,
our model has only two outputs, which are EXPR and VA, since the last
training task reuses these two outputs for computing loss.

\subsection{Multitask training with missing labels}
\label{sec:multitask_missing_labels}
Besides the excluding of facial AU detection task, we also would like
to take into account one important aspect of the Affwild2  
dataset, which is the fact that there are 164 videos that are being
annotated with both VA and EXPR. Instead of treating   
all the training videos as if they are annotated with only one label
like \cite{Deng}, we check if the given video frame has  
annotated with one or both EXPR and VA labels. Then, we compute the
objective loss of the secondary task using the distillation loss alone
or supervision loss plus distillation loss, respectively.

Let $(X, Y)$ be the training dataset, where $X$ is a set of input
vectors and $Y$ is a set of ground truth training labels. 
Since our dataset contains three parts including: \textit{Mixed EXPR}, \textit{Mixed VA} and \textit{Affect EXPR\_VA}, 
therefore $(X, Y) = \{(X^{(i)}, Y^{(i)})\}^3_{i=1}$. For convenience
of notation, we assume each subset $i$ includes an equal number $N$  
of instances within a batch, i.e $(X^{(i)}, Y^{(j)}) = \{(x^{(i, n)}, y^{(i, n)})\}^N_{n=1}$ where $n$ indexes the instance.
Because the data from the last set \textit{Affect EXPR\_VA} is including both 
EXPR and VA annotations, we denote $3_{expr}$ and $3_{va}$ as the EXPR
annotation and the VA annotation of this set, respectively.  
For example, instance $x^{(3, 1)}$ belongs to \textit{Affect EXPR\_VA} dataset and has two annotations: $y^{(3_{expr}, 1)}$ and $y^{(3_{va}, 1)}$

The inputs for all instances have the same dimensionality, regardless of task. 
However, the ground truth labels for different tasks have different dimensionality.  The label for the first task
(EXPR) is $y^{(1)} \in \{0,1\}^7$. The label for the second task (VA)
is $y^{(2)} \in [-1,1]^2$. The label for the last task (EXPR\_VA) is the mixed of the two tasks above.

\begin{figure}[ttb]
	\centering
	\includegraphics[width=.9\linewidth]{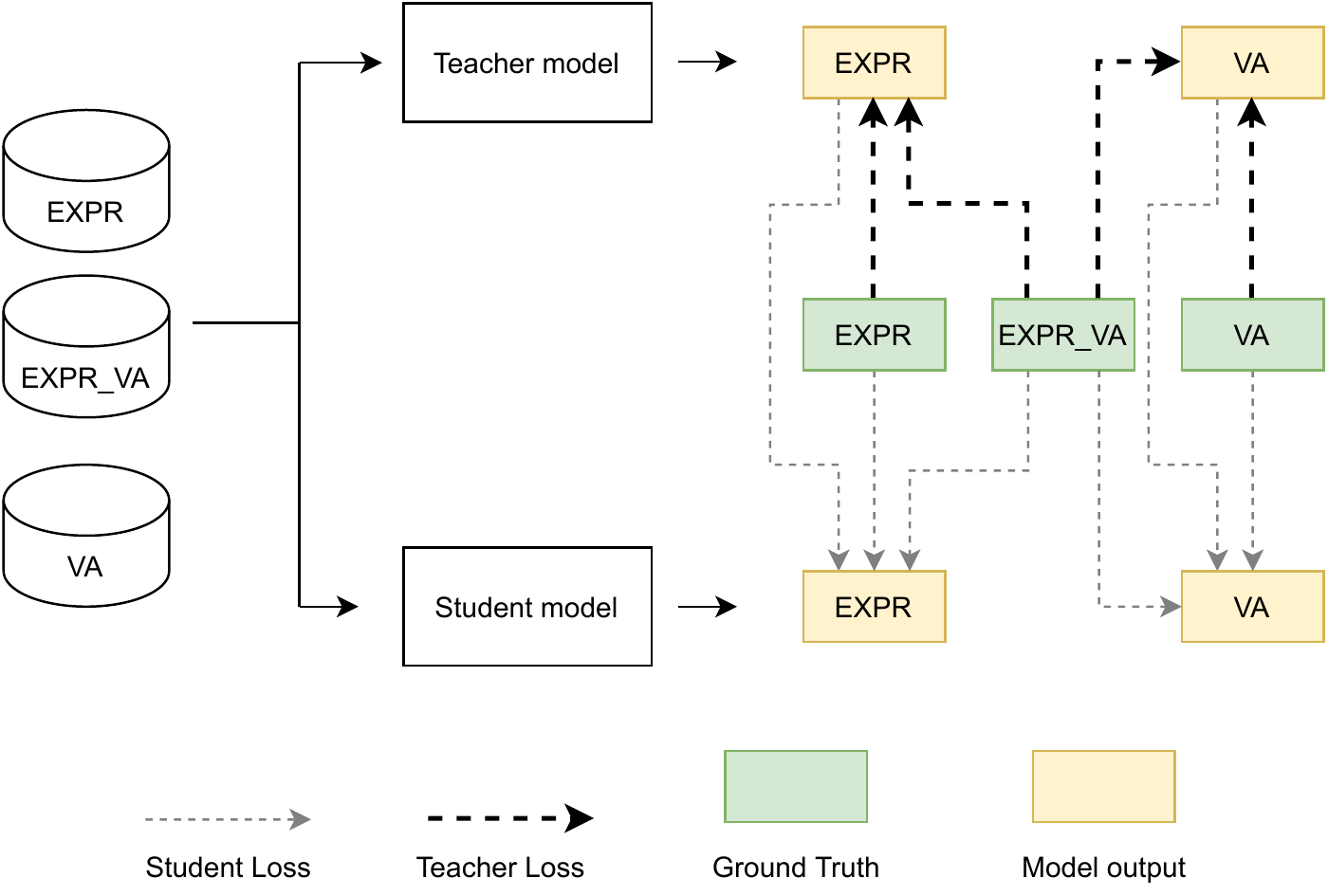}
	\caption{The overview of our multitask training with missing labels. }
	\label{fig:multitask}
\end{figure}

\begin{figure}[ttb]
	\centering
	\includegraphics[width=.5\linewidth]{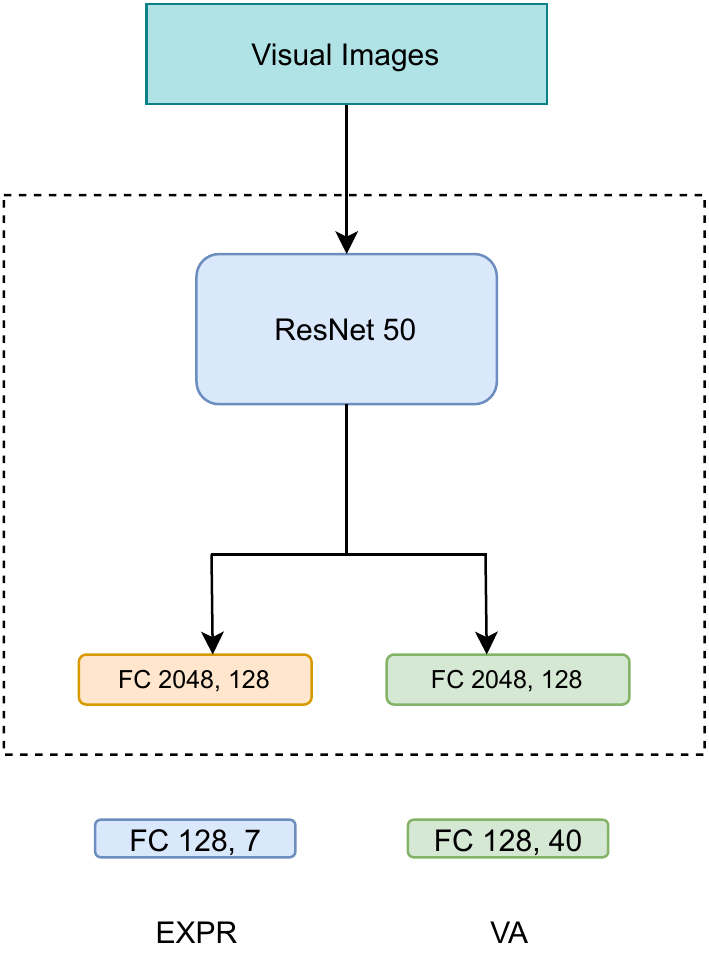}
	\caption{The model architecture of our networks. }
	\label{fig:abaw_architecture}
\end{figure}

Similar to \cite{Deng}, we also apply the two step training for
capturing inter-task correlations. We train a single teacher model  
using only the ground truth labels in the first step. In the second step, we replace the missing labels with soft 
labels derived from the output of the teacher model. We then use the
ground truth and soft labels to train a single student model. 
Different from  \cite{Deng}, we do not train multi student models for
model ensemble because this approach is too costly  
in term of computation. We believe that by applying an appropriate
training method, the model will be able to reach the same or even  
higher in performance, comparing to the ensemble approach. The overview of our network can be seen in Fig \ref{fig:multitask} and the architecture of
our model is in Fig \ref{fig:abaw_architecture}.

To be in the same line with \cite{Deng} in the sense of notation, we also denote the output of our multitask network by 
$f^{(i)}_\theta(\cdot)$ where $\theta$ contains the model parameters of either teacher model 
or student model, and $i \in \{1, 2\}$ indicates the current task. For example, $f^{(1)}_{\theta}(x^{(3)}$ indicates
the output of the network for task 1 (EXPR) for an instance in the \textit{Affect EXPR\_VA} set. To avoid clutter,
we will often refer to the output of the teacher network on task $i$
by $t^{(i)}$ irrespective of what the input label is, 
i.e. $t^{(i)} = f_\theta^{(i)}(x^{(j)})$ for some $j \in \{1, 2, 3\}$ and similarly to the output of the student network 
on task $i$ by $s^{(i)}$. 

Regarding the objective loss functions, similar to \cite{Deng}, we
also treat the problem of expression classification as  
a multiclass classification problem, and the problem of valence-arousal estimation as a combination of multiclass 
classification and regression problem. We will use the same soft-max function $SF$, the cross entropy 
function $CE$ and the concordance correlation coefficient function $CCC$, which have already been defined in \cite{Deng}.

\subsubsection{Supervision loss functions} Here we denote the loss functions that are used for optimizing our models 
parameters with the supervision of the ground truth labels for each of our training tasks.
\paragraph{EXPR task} The supervision loss for the samples from the \textit{Mixed EXPR} set is denoted as:
\begin{equation}
	\mathscr{L}^{(1)}(y^{(1)}, t^{(1)}) = CE\left(y^{(1)}, SF(t^{(1)}, 1)\right)
\end{equation}

\paragraph{VA task} The supervision loss for the samples from the \textit{Mixed VA} set is denoted as:
\begin{multline}
	\label{eq:va_loss}
	\begin{split}
	\mathscr{L}^{(2)}(y^{(2)}, t^{(2)}) & = \sum^2_{i=1} \Biggl\{ CE\left(onehot(y^{(2)}_i), SF(t^{(2)}_i, 1)\right) \\
	&\qquad\qquad+ \frac{1}{B}\left(1 - CCC(y_i^{(2)}, t_i^{(2)})\right) \Biggr\}
	\end{split}
\end{multline}

\paragraph{EXPR\_VA task} For the samples from Affect EXPR\_VA set, since this set is annotated using both VA and EXPR, 
the supervision loss for this task is denoted as:
\begin{multline}
	\begin{split}
	\mathscr{L}^{(3)}(y^{(3)}, t^{(3)}) & = CE\left(y^{(3_{expr})}, SF(f^{(1)}_{\theta_t}(x^{(3)}), 1)\right) \\
	&+ \sum^2_{i=1} \Biggl\{ CE\left(onehot(y_i^{(3_{va})}), SF(f^{(2)}_{\theta_ti}(x^{(3)}), 1)\right)	\\
	&\qquad\qquad+ \frac{1}{B}\left(1 - CCC(y_i^{(3_{va})}, f^{(2)}_{\theta_ti}(x^{(3)}))\right) \Biggr\}
	\label{eq:expr_va}
	\end{split}
\end{multline}


\subsubsection{Distillation loss functions} 
\label{sec:distillation_loss}
Here we denote the loss functions that are used to optimize our student model parameters 
with the supervision of both the ground truth labels (hard targets) and the pretrained teacher model's outputs (soft targets)
 for each of our training tasks.
Similar to \cite{Deng}, we use the $KL$ divergence $KL(p, q) = \sum_ip_ilog\left(\frac{p_i}{q_i}\right)$ 
to measure the different between two probability distributions 
(output of teacher model and student model).


\paragraph{EXPR task} Distillation loss for the samples from the \textit{Mixed EXPR} set:
\begin{align}
	\mathscr{H}^{(1)}(t^{(1)}, s^{(1)}) = KL\left(SF(t^{(1)}, T), SF(s^{(1)}, T)\right) 
\end{align}

\paragraph{VA task} Distillation loss for the samples from the \textit{Mixed VA} set:
\begin{align}
	\mathscr{H}^{(2)}(t^{(2)}, s^{(2)}) = \sum^2_{i=1} KL\left(SF(t_i^{(2)}, T), SF(s_i^{(2)}, T)\right) 
\end{align}

\paragraph{EXPR\_VA task} Distillation loss for the samples from the \textit{Affect EXPR\_VA} set is the 
combination of the EXPR and VA distillation losses, which is denoted as:
\begin{align}
	\begin{split}
	\mathscr{H}^{(3)}(t^{(3)}, s^{(3)}) = 
	KL\biggl(SF\left(f^{(1)}_{\theta_t}(x^{(3)}), T\right), SF\left(f^{(1)}_{\theta_s}(x^{(3)}), T\right)\biggr) 
	\\
	+ \sum^2_{i=1} KL\biggl(SF\left(f^{(2)}_{\theta_ti}(x^{(3)}), T\right), SF\left(f^{(2)}_{\theta_si}(x^{(3)}), T\right)\biggr) 
	\end{split}
\end{align}

\subsubsection{Batch-wise loss functions}
Given a batch of data $\left(X, Y\right) = \{\{(x^{(i, n)}, y^{(i, n)})\}^N_{n=1}\}^3_{i=1}$, 
the parameters of teacher network and student networks are denoted as $\theta_t$ and $\theta_s$, 
respectively. Since our last dataset \textit{Affect EXPR\_VA} contains annotation for both EXPR and VA,
therefore, when $i = 3$ then $y^{(3, n)}$ contains both $y^{(3_{expr}, n)}$ and $y^{(3_{va}, n)}$.

The training teacher loss:
\begin{equation}
	\mathscr{F}_t(X, Y, \theta_t) = \sum_{i=1}^3\sum_{n=1}^N\mathscr{L}^{(i)}\left(y^{(i, n)}, f_{\theta_t}^{(i)}(x^{(i, n)})\right)
	\label{eq:teacher}
\end{equation}

The loss of sample $x$ with ground truth $y$ from dataset $i$ with $i \in \{1, 2, 3\}$ will be denote as:
\begin{multline}
	\label{eq:loss_x}
	\begin{split}
	\mathscr{G}_i(x, y, \theta_t, \theta_s) & = 
 		\lambda \times \mathscr{L}^{(i)}\left(y, f_{\theta_s}^{(i)}(x)\right) \\
		&\qquad + \; (1 - \lambda) \times \mathscr{H}^{(i)}\left(f^{(i)}_{\theta_t}(x, f_{\theta_s}^{(i)}(x)\right)
	\end{split}
\end{multline}

Similar to \cite{Deng}, we also use the parameter $\lambda$ to weight the supervision loss versus the distillation loss.
The $\lambda$ parameter is set to $0.6$ to weight the ground truth slightly more than the soft labels.

The student loss is denoted as:
\begin{align}
	\begin{split}
	\mathscr{F}_t(X, Y, \theta_t, \theta_s)	& = \sum_{n=1}^N \mathscr{G}_3\left(x^{(3, n)}, y^{(3, n)}, \theta_t, \theta_s\right) \\
	&\;\;\; + \sum_{i=1}^2\sum_{n=1}^N \Bigg\{ \mathscr{G}_i\left(x^{(i, n)}, y^{(i, n)}, \theta_t, \theta_s\right) \\
	&\qquad\;\;\; + \sum_{j \neq i} \mathscr{H}^{(j)} \left( f^{(j)}_{\theta_t} (x^{(j, n)}), f^{(j)}_{\theta_s} (x^{(j, n)}) \right)
	\Bigg\}
	\end{split}
	\label{eq:student}
\end{align}

As we have mentioned earlier, there are 164 videos that are annotated with both EXPR and VA in the Affwild2 database. 
To exploit these sharing annotations, we propose a method to compute student loss with taking into account this 
characteristic, which is denoted as:

\begin{multline}
	\begin{split}
	&\mathscr{F}_t(X, Y, \theta_t, \theta_s)	= \sum_{n=1}^N \mathscr{G}_3\left(x^{(3, n)}, y^{(3, n)}, \theta_t, \theta_s\right) \\
	&+ \sum_{i=1}^2\sum_{n=1}^N 
	\Bigg\{ \mathscr{G}_i\left(x^{(i, n)}, y^{(i, n)}, \theta_t, \theta_s\right)	\\
	&\qquad+ \sum_{j \neq i} 
		\begin{cases}
			\mathscr{H}^{(j)} \left( f^{(j)}_{\theta_t} (x^{(j, n)}), f^{(j)}_{\theta_s} (x^{(j, n)}) \right),& \text{if } y^{j, n} \text{ is NA}\\
			\mathscr{G}_j\left(x^{(j, n)}, y^{(j, n)}, \theta_t, \theta_s\right), & \text{otherwise}
		\end{cases}
	\Bigg\}
	\Bigg\}
	\end{split}
	\label{eq:student_if}
\end{multline}

As described in Equation \ref{eq:student_if}, we can see that for each sample, instead of treating it as having only 
one label, we check if it contains a secondary label or not, then compute the distillation loss only or 
supervision loss plus distillation loss for this secondary label, correspondingly.

\subsection{Visual images analysis}
\label{sec:visual}
For the visual images, face images with size of $height \times width$ pixels are aligned and extracted from each video frame. Then, we use 
these images to train a ResNet 50 model using the method mentioned in Section \ref{sec:multitask_missing_labels}. During training, we have 
applied some image-wise augmentation filters to improve the performance of the model. These filters are 
including: random image translation \cite{electronics9111892} and random image horizontal flip.

\subsection{Temporal information exploiting using GRU network}
Once the student model of the ResNet 50 network have been trained, we use this model to extract features 
from each video frame. Then, we group these features together to form a new dataset $ds$ of feature's sequences
with sequence length of $32$ frame per sequence. Finally, we fed data from this new dataset $ds$ into a bidirectional GRU network for exploiting temporal information, as well as 
performing seven basic emotions classification and valence-arousal estimation. Regarding the GRU model's parameters, we also use the training method in 
Section \ref{sec:multitask_missing_labels} to train this model's parameters. During the training, we have used 
the same augmentation filters that are mentioned in Section \ref{sec:visual} but in sequence level.

\section{Experiments and Results}
\subsection{Implementation details}
The whole network system is implemented using PyTorch framework \cite{NEURIPS2019_9015}. During the training phase, Adam optimizer \cite{kingma2017adam} were employed with 
the initial learning rate is set of $1e^{-4}$. 
The maximum number of epochs is 40 and the training process will stop when there is no improvement after five consecutive epochs. The number of batch size for the 
CNN part of the network is set to 64. For RNN network, the batch size is 16.
The training and validating processes were performed on an Intel Workstation machine 
with a NVIDIA Gerforce RTX 2080 Ti 11G GPU.
\subsection{Results}
Here we report the results of different experiments to demonstrate the effectiveness of each of our changes compare to the original method \cite{Deng}. For the evaluation metrics,
we use the same criterion as outlined in \cite{kollias2020analysing}. Valence and Arousal estimation is based on the mean Concordance Correlation Coefficient (CCC).
The seven basic expressions classification is measured by $0.67 \times F_1 \;\text{score} + 0.33 \times \text{total accuracy}$. For each of our experiments, we run it 10 times and
report the mean of the evaluation results on the Validation set of the AffWild2 dataset.

Table \ref{tbl:teacher_cnn} shows the performance of the teacher network when training using Equation \ref{eq:teacher} 
with only the first two tasks ($\mathscr{T} \in \{1, 2\}$) and 
with all three tasks ($\mathscr{T} \in \{1, 2, 3\}$). From this table, we can see that when training with only two tasks, 
our model has already outperformed the baseline results of the competition. This finding is in the same line with \cite{Deng}, which 
is the proof of the effectiveness of data balancing and multitask training method.
Now, when we add the third task EXPR\_VA into the training process
($\mathscr{T} \in \{1, 2, 3\}$), we can see that the value of both EXPR and Valence have increased quite a lot, especiately the later with 17\% of improvement.
Despite of having a slightly decreasing in term of Arousal (about 2\%), the performance of the network has improved 
in overall by a large margin, compared to the model trained without the EXPR\_VA task. 

\begin{table}[hbt]
	\caption{Performance results of the teacher CNN models on the validation set of the Affwild2 database.}
	\begin{center}
	\begin{tabular}{|l|c|c|c|c|}
	\hline
	Method & EXPR & Valence & Arousal \\
	\hline
	Baseline & 0.366 & 0.23 & 0.21 \\ 
	\hline
	Multitask $\mathscr{T} \in \{1, 2\}$ & 0.498 & 0.374 & \textbf{0.407} \\ 
	\hline
	Multitask $\mathscr{T} \in \{1, 2, 3\}$ & \textbf{0.513} & \textbf{0.438} & 0.398 \\ 
	\hline
	\end{tabular}
	\end{center}
	\label{tbl:teacher_cnn}
	\end{table}

After training the teacher model, we train student models with the supervision of both ground truth and the 
pretrained teacher model using Equation \ref{eq:student} for the case of not using the shared annotations (No sharing), and using Equation 
\ref{eq:student_if} for the case of using the shared annotations (With sharing). The results are showed in 
Table \ref{tbl:student_cnn}. From this table, it can be seen that the performance of the model trained using 
the shared annotations (With sharing) is better than the one trained without using it (No sharing). This results 
indicate the important of exploiting the sharing annotations in the database.
\begin{table}[hbt]
	\caption{Performance results of the student CNN models on the validation set of the Affwild2 database. 
	The student models are trained using all three tasks $\mathscr{T} \in \{1, 2, 3\}$. 
	}
	\begin{center}
	\begin{tabular}{|p{3.2cm}|c|c|c|c|}
	\hline
	Method & EXPR & Valence & Arousal \\
	\hline
	No sharing & 0.513 & \textbf{0.472} & 0.412 \\ 
	\hline
	With sharing & \textbf{0.525} & 0.471 & \textbf{0.421} \\ 
	\hline
	\end{tabular}
	\end{center}
	\label{tbl:student_cnn}
	\end{table}

Once the student model is trained, we use this CNN model to extract
features to train GRU network for exploiting temporal  
information. We train a teacher model using Equation \ref{eq:teacher}
and a student model using Equation \ref{eq:student_if}. 
Table \ref{tbl:gru} shows the results of these models. From this
table, we can see that: the performance of student model  
is better the teacher model in all cases. And when comparing with the CNN model (in Table \ref{tbl:student_cnn}), 
the CNN + GRU model outperformed it by a large margin.
\begin{table}[hbt]
	\caption{Performance results of the CNN + GRU model. Both teacher and student models are trained using all three tasks 
	$\mathscr{T} \in \{1, 2, 3\}$. 
	}
	\begin{center}
	\begin{tabular}{|p{3.2cm}|c|c|c|c|}
	\hline
	Method & EXPR & Valence & Arousal \\
	\hline
	Teacher model & 0.555 & 0.523 & 0.543 \\ 
	\hline
	Student model & \textbf{0.555} & \textbf{0.526} & \textbf{0.551} \\ 
	\hline
	\end{tabular}
	\end{center}
	\label{tbl:gru}
	\end{table}

\subsection{Comparison with State of the art}

Here we compare the performance of our model with the state of the art on the validation set of Affwild2 dataset.
Although in this 2nd challenge, the database has been updated by adding more videos and labels for the 
AU detection task. But since the data for seven basic expression
detection task (EXPR) and valence-arousal estimation task 
(VA) are almost unchanged, we are still able to compare the performance of our model with the works on the previous challenge.
\begin{table}[hbt]
	\caption{Comparison with other works on the validation set of the Affwild2 database}
	\begin{center}
	\begin{tabular}{|p{3.2cm}|c|c|c|c|}
	\hline
	Method & EXPR & Valence & Arousal \\
	\hline
	Zhang et al. \cite{zhang2020m3t} & - & 0.32 & 0.55 \\ 
	\hline
	Gera et al. \cite{gera2020affect} & 0.465 & - & - \\ 
	\hline
	Deng et al. \cite{Deng} & 0.493 & 0.335 & 0.515 \\ 
	\hline
	Kuhnke et al. \cite{Kuhnke_2020} & 0.546 & 0.493 & \textbf{0.613} \\ 
	\hline
	Our model & \textbf{0.555} & \textbf{0.526} & 0.551 \\ 
	\hline
	\end{tabular}
	\end{center}
	\label{tbl:stoa}
 	\end{table}

Table \ref{tbl:stoa} shows the comparison results between the works on the same dataset, which including 
Zhang et al. \cite{zhang2020m3t} with their $M^3T$ model, Gera et al. \cite{gera2020affect} with spatio-channel attention 
network, Deng et al. \cite{Deng} with their multitask model trained on multiple datasets with incomplete labels, Kuhnke 
et al. \cite{Kuhnke_2020} with their two streams aural-visual network. From this table, we can see that our model 
outperformed other works in term of EXPR (seven basic facial expression classification) and Valence estimation. Despite of 
having a decreasing in term of Arousal estimation, our model is still in comparable with the state of the art.
In term of speed, Kuhnke et al. \cite{Kuhnke_2020} method is using 3D CNN network, which is known to 
have the issues of time-consumption, training complexities and hardware memory usage 
\cite{kumawat2019lp3dcnn, kang2021efficient, 9362857}, our method seems to have less 
complexity and time-consuming compared to their method. When we 
compare our results with the work of Deng et al. \cite{Deng}, which is the most closed work compared to us, we can see 
that our model is outperformed their model ensemble in all tasks, in the same time our model is five times faster compared
to their model because their model are the combination of five models ensemble.

\section{Conclusion}
In this paper, we have presented a method to optimize multitask training with imcomplete labels. By adding a new task 
to train deep neural network on a dataset which contains both seven
basic expressions and valence-arousal values, along with exploiting  
the shared annotations inside the Affwild2 database when training student model, resulting a model that is better than 
state of the art in term of seven basic expression classification and valence estimation on the validation set of the 
Affwild2 database.
In future work, we will investigate about multimodal network, e.g. multitask visual-aural neural network for analyzing 
both visual and aural streams from a video recording.



\normalsize
\bibliography{references}


\end{document}